\documentclass[11pt,a4paper]{article}
\usepackage[hyperref]{naaclhlt2019}
\usepackage{times}
\usepackage{latexsym}

\usepackage{url}

\usepackage{xcolor}

\usepackage{color}
\usepackage{multirow}
\usepackage[normalem]{ulem}
\usepackage{subcaption}
\usepackage{algorithmic}
\usepackage{amsmath,amsfonts,amssymb}
\usepackage{mathtools}
\usepackage{booktabs}
\usepackage{acronym}
\usepackage{mathtools}

\usepackage{mathtools}
\usepackage{amsmath}

\DeclareMathOperator*{\argmin}{arg\,min}
\usepackage{microtype}

\aclfinalcopy 

\title{Beam Search with Bidirectional Strategies for Neural Response Generation}

\author{Pierre Colombo$^{\dag\star}$, Chouchang (Jack) Yang$^*$,  Giovanna Varni$^\star$, Chlo\'e Clavel$^\star$ \\
  $^\star$T\'el\'ecom ParisTech, Universit\'e Paris Saclay \\
  $\dag$ IBM GBS France\\
  $^*$ University of Washington \\
  \texttt{pierre.colombo@ibm.com} \\
    \texttt{ccjack@uw.edu} \\
    \texttt{giovanna.varni@telecom-paris.fr} \\
  \texttt{chloe.clavel@telecom-paris.fr} \\
  }

\date{}

\begin{document}

\maketitle

\begin{abstract}
Sequence-to-sequence neural networks have been widely used in language-based applications as they have flexible capabilities to learn various language models. However, when seeking for the optimal language response through trained neural networks, current existing approaches such as beam-search decoder strategies are still not able reaching to promising performances. Instead of developing various decoder strategies based on a ``regular sentence order" neural network (a trained model by outputting sentences from left-to-right order), we leveraged ``reverse'' order as additional language model (a trained model by outputting sentences from right-to-left order) which can provide different perspectives for the path finding problems. In this paper, we propose bidirectional strategies in searching paths by combining two networks (left-to-right and right-to-left language models) making a bidirectional beam search possible. Besides, our solution allows us using any similarity measure in our sentence selection criterion. Our approaches demonstrate better performance compared to the unidirectional beam search strategy.

\end{abstract}

\section{Introduction}
\label{introduction}
Seq2seq models have shown state-of-the-art performance in tasks such as machine translation \cite{Sutskever}, neural conversation \cite{nlp2015}, image captioning \cite{Venugopalan_2015_ICCV}, and text summarization \cite{DBLP}, exhibiting human-level performance \cite{human}. Despite some drawbacks (e.g huge parallel corpora are needed to train seq2seq models, expert knowledge required to set the hyperparameters), seq2seq models are becoming increasingly popular and are now deployed in real world applications \cite{real_product}.
During inference, a trained seq2seq model aims to find the best sentence given a source sentence. Since searching for all the possible paths is not practical but also computationally expensive, existing work relies on beam search based algorithms to solve this issue \cite{Lipton2015ACR}. Current solutions have limited performances due to three major constraints: (1) beam search sentence selection is based on likelihood regardless of the evaluation metric,  (2) during the generation process, only left to right dependencies (right to left are ignored) are considered by the model, (3) seq2seq strongly foster safe sentences \cite{mmi}: during generation, the influence of the input decreases while words are generated \cite{intro} meaning that the end of the sequence is less likely to be input relevant. Those limitations constraint beam search performances: on Switchboard Corpus with a Beam Size of 50 optimal re-ranking would yield to an improvement of 128\% (see Supplementary for full table).
For the evaluation metric we follow \citet{mmi,bleu_choice_2} adopt the BLEU-4 score to compare the algorithms.
\newline In this work, we introduce two novel algorithms based on beam search (\textit{VBS}) with different ranking procedure: (1) \textit{BidiS} a generalisation of the work of \citet{overgenerate,forward_back} that uses a "reverse" decoder to re-score the produced sentence penaliszing sentences whose end is less likely given the input, (2) \textit{BidiA} an algorithm, that looks at the closest pair by incorporating the similarity measure between two beams of hypothesis; one with sentences generated in the regular order, the other with sentences generated in the reverse order. Our results show that leveraging the reverse order can boost beam search performance leading to higher BLEU-4 score and more diverse responses compared to \textit{VBS}. Complexity analysis further shows that our proposed algorithms have dramatically reduced computational cost compared to the traditional approaches.

\section{Models}
\label{models}
We notice that limitation (2) and (3) previously introduced can be solved by introducing bidirectionally in the beam search. Indeed, training a seq2seq to output the sentence in the reverse order can model right to left dependencies and reduces the path length between the input and the end of the sentence (the shorter the paths are, the easier it is to model dependencies) making the end of the sentence more dependent of the input and producing more diverse sequences and model right to left dependencies as well.

\subsection{Preliminaries} 
\textbf{Vanilla Beam Search} (\textit{VBS})
We denote B as the beam size, T as the maximum sentence length and V as the vocabulary size. A RNN encoder takes an input sequence $X = (x_1, ..., x_T)$, to learn the language model word by word during the training phase. When a language task is given, the decoder travels through paths at each time and keeps the $B$ most likely sequences. At each step, \textit{VBS} considers at most $V \times B$ hypothesis. The sequence likelihood is measured by using the score function in \cite{Wu2016GooglesNM}
\begin{equation}
    s(Y,X)=\frac{\log P(Y|X) }{lp(Y)}
\end{equation}

where X is the source, Y is the current target, and $lp(Y)=\frac{(5+|Y|)^\alpha}{(5+1)^\alpha}$ is the length normalization factor. We select $\alpha = 0.6$, which produce a higher BLEU score as illustrated in \cite{Wu2016GooglesNM}. The beam search is stopped when exactly $B$ finished candidates have been found \cite{attention}. In the worst case, the algorithm will run for a maximum of $T$ steps.

\textbf{Regular and Reverse model training}
For the bidirectional beam search we train two different networks over the same dataset. The first seq2seq network called "regular" is trained to predict the sentence in the regular order. The second network called "reverse" is trained to predict the sentence in the reversed order. For example, if the regular network is trained with the pair ``What do you like ?" / ``I like cats !", the reversed is trained with the pair ``What do you like ?" / ``! cats like I". During decoding, the reverse model estimates right-to-left dependencies while the regular model estimates left-to-right dependencies. \footnote{From graph topology viewpoints, the decoder procedure from the right side is very different from the left sides. During exploring graph from left to right the regular seq2seq faces a huge very likely first token while the reverse seq2seq has a very restraint choice (mainly punctuation). More details are included in Supplementary.} Two different settings have been explored: (1) training two independent seq2seq, (2) sharing the encoder of the two seq2seq and training using the following loss $\mathcal{L}$ where: 
\begin{equation}
    \mathcal{L} = \alpha \mathcal{L}_{Rg} + (1 - \alpha ) *\mathcal{L}_{Re}
\end{equation}\label{eq:multitask}
$\mathcal{L}_{Rg}$ is the Cross Entropy loss computed with the regular decoder and $\mathcal{L}_{Re}$ is the Cross Entropy loss computed with the reverse decoder. Since both approaches exhibit comparable performances, we choose to share the encoder to minimise the number of parameters in our model.

\subsection{Beam Search with Bidirectional Scoring (\textit{BidiS})}
A Beam search generates word by word from left to right: the token generated at time step $t$ only depending on past token, but would not affected by the future tokens. Inspired by the work of \cite{mmi}, we propose a Beam Search with Bidirectional Scoring (\textit{BidiS}), which scores the $B$ best candidates generated by the regular seq2seq model as follows:
\begin{equation}
    \resizebox{.89\hsize}{!}{$s(Y_T,X)=\frac{\log P(Y_T^+|X) }{lp(Y_T)} + \lambda \times \frac{\log P(Y_T^-|X) }{lp(Y_T)}$}
\end{equation}
where $Y_T^+$ and $Y_T^-$ represents the final sequence in the regular order and reversed order respectively. Moreover, $P(Y_T^+|X)$ is computed by using the regular model while $P(Y_T^-|X)$ is computed by using the reverse model.  $\lambda$\footnote{Optimisation process shows that $\lambda$ compensate the difference of scale between $\frac{\log P(Y_T^+|X)}{lp(Y_T)}$ and $\frac{\log P(Y_T^-|X) }{lp(Y_T)}$} is optimized in the validation. 
The intuition here is as follows: after generation of $B$ sentences from the regular seq2seq, the reverse model computes $P(Y_T^-|X) $, and assigns higher probabilities to sequences presenting a more likely right-to-left structure and more likely ending given the input.  Since $B$ best lists produced by our models are grammatically correct, the final selected options are well-formed and present the best combination of both directions.

\subsection{Beam Search with bidirectional agreement (\textit{BidiA})}
The previous algorithm has two weaknesses. Firstly it introduces a hyperparameter $\lambda$. Secondly, the reverse model is only used to re-score the sentences generated by the regular model, meaning that potentially good sentences generated by the reverse model are not considered. We solve these two problems by proposing a Beam Search with bidirectional agreement (\textit{BidiA}); a hyperparameter free algorithm that uses $\frac{B}{2}$ best sequences according to the reverse seq2seq model. Formally if $S$ and $S\prime$ are the sets containing $\frac{B}{2}$ sequence generated by the regular and reverse model respectively, we output $Y_{n_0}$ such that: 

\begin{equation}
    (Y_{n_0},Y_{r_0})  = \argmin_{Y_n \in S,Y_r \in S\prime }{1 - sim(Y_n,Y_r)}
\end{equation}

where $sim$ represents any similarity measure between two sentences \footnote{$sim$ does not need to be differentiable.}. For our experiment, we propose two different choices: (1) an adaptation of the BLEU score, $BLEU_T$ where the corpus length is set to $T$ to foster longer responses formally:
\begin{equation}
   BLEU_T = BP_T \times exp(\sum_{n=1}^N w_n log(p_n))
\end{equation}
where the brevity penalty is set to $BP_T = \text{exp}(1 - \frac{T}{c})$\footnote{Brevity penalty introduces diversity and foster longer sentences.}, $p_n$ is the geometric average of the modified n-gram precisions, using n-grams up to
length N and $w_n$ are positive weights summing to one, (2) an adaptation of the Word Mover's Distance ($WMD_T$) \cite{mover_distance} (stopwords are removed and final score is multiplied by $BP_T$) that captures the relationship between words, by computing the “transportation” from one phrase to another conveyed by each word \footnote{Implementation details are given in supplementary}.

\section{Results}
\label{tasks}

\subsection{Corpora and Metrics}
\textbf{Corpora}: We evaluate our algorithms on two spoken datasets (specific phenomena appear when working with spoken language \cite{dinkar2020importance} compared to written text). (1) The Switchboard Dialogue Act Corpus (SwDA)a telephone speech corpus \cite{swa}, consisting of about 2.400 two-sided telephone conversation. (2) The Cornell Movie Corpus \cite{cornell} which contains around 10K movie characters and around 220K dialogues. \newline
\textbf{Metrics}: To evaluate the performance and language response quality for each decoder strategy, we use two classical different metrics at the sentence level. (1) A BLEU-4 score \cite{bleu}  is computed on the unigrams, bigrams, trigrams and four-grams; and then micro-averaged. (2) A Diversity score: distinct-n \cite{mmi} is defined as the number of distinct n-grams divided by the total number of generated words. Indeed, in neural response generation, we want to avoid generate generic responses such as "I don't know", "Yes", "No" and foster meaningful responses.

\subsection{Response Quality}

\begin{figure*}[!htb]

\minipage{0.5\textwidth}
\centering
  \includegraphics[width=0.65\linewidth]{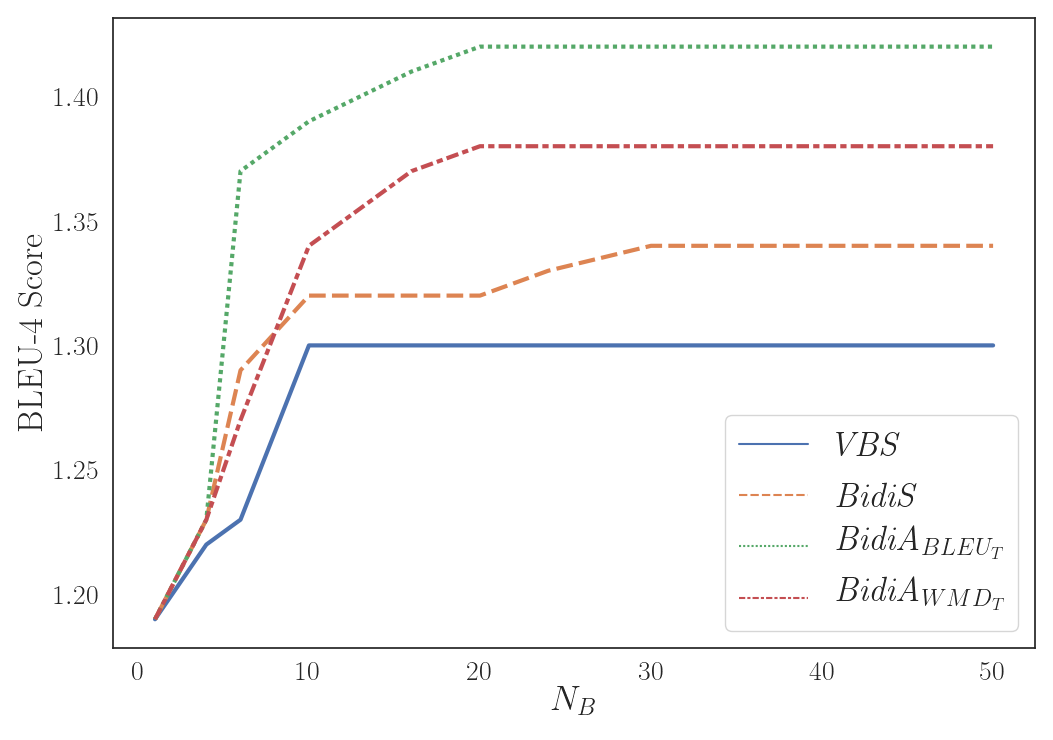}
\endminipage\hfill
\minipage{0.5\textwidth}
\centering
    \includegraphics[width=0.65\linewidth]{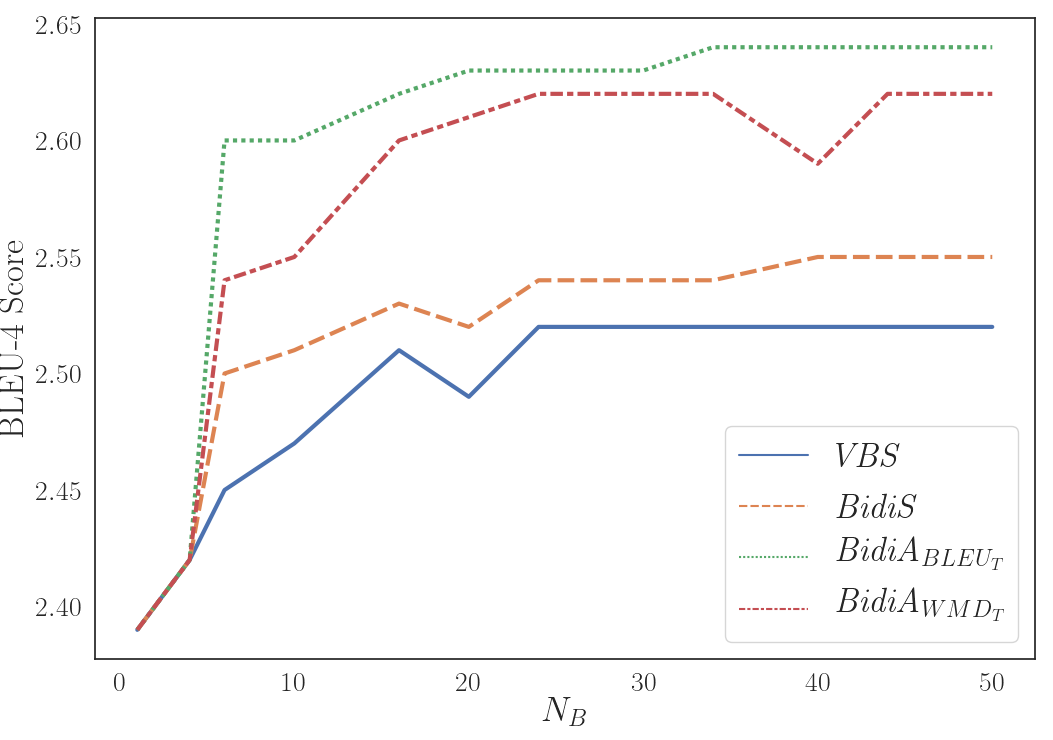}
\endminipage\hfill
\minipage{0.33\textwidth}%

\endminipage
   \caption{\textbf{BLEU-4 Scores} for the proposed algorithms on two different datasets: Cornell (left) and SWA (right). $N_B$ is the beam size for \textit{VBS} and \textit{BidiS} and 2 times the beam size for $\textit{BidiA}_{WMD_T}$ and $\textit{BidiA}_{BLEU_T}$}
   \label{fig:bleues}\vspace{-0.5cm}
\end{figure*}

\autoref{fig:bleues} shows our proposed system results in BLEU-4 score metric. We see that our proposed methods (\textit{BidiS} and \textit{BidiA}) achieve better performances than \textit{VBS} showing that bidirectionality boosts performances. $\textit{BidiA}_{BLEU_T}$ achieves the best result overall yielding to a relative improvement of 9\% on Cornell and 5\% on SWA. From \autoref{fig:bleues}, we see that for two different metric $sim$ \textit{BidiA} leads to better results than both other algorithm. Improvement of \textit{BidiS} over the baseline \textit{VBS} shows that the optimisation of $\lambda$ on the validation set leads to good generalisation on the test set. $\textit{BidiA}_{BLEU_T}$ is slightly better than $\textit{BidiA}_{WMD_T}$ which is likely to be related to the choice of evaluation metric. 
From \autoref{fig:bleues} we can see that the BLEU-4 score of \textit{VBS} stop increasing when $B>10$. \textit{BidiS} and \textit{BidiA} keep improving the quality of the sequence while more hypothesis are proposed. This suggests that our bidirectional beam search is more efficient at extracting best sentence as the number of hypothesis increases.
From \autoref{fig:bleues} we can see that \textit{VBS}, \textit{BidiA} and $\textit{BidiA}_{WMD_T}$ present a drop in the performance for a number of hypothesis of 20 and 40: when performing the beam search for 20 hypothesis we observe that the seq2seq is very confident about sentences that lead to lower BLEU-4 score. Those sentences are not considered when $B<20$ and better sentences are extracted when the beam size increases. $\textit{BidiA}_{BLEU_T}$ does not present this drop of performance this is due to the metric choice (based on overlaps) that selected different sentences from $\textit{BidiA}_{WMD_T}$.

\subsection{Rank Analysis}
\begin{figure}
    \centering
    \includegraphics[width=0.4\textwidth]{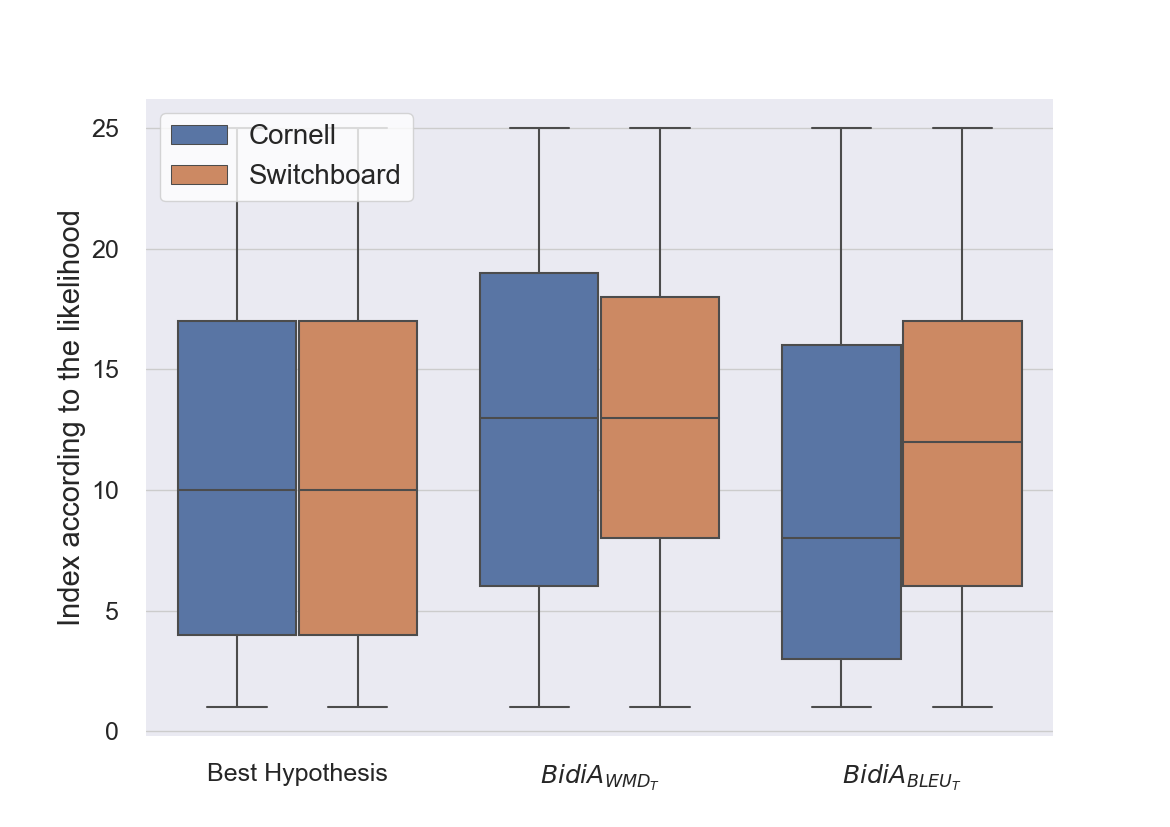}
  \caption{\textbf{Index of the response}. Index is the position of the sentence in the beam returned by \textit{VBS}: the most likely sequence is ranked 1, the less likely is ranked 25. \textit{Best Hypothesis} is the sentence (hypothesis) in the beam that yields to the highest BLEU-4.}
  \label{fig:rank_analysis}\vspace{-0.5cm}
\end{figure}

In this section we compare the index returned by \textit{BidiA} and \textit{Best Hypothesis} as shown in \autoref{fig:rank_analysis}. \autoref{fig:rank_analysis} illustrates one of the limitation of likelihood based ranking when an off-shell metric is used for evaluation: the very most likely sentences are not the one with the highest BLEU-4.  Interestingly, index distribution of \textit{Best Hypothesis} is very similar for both Cornell and Switchboard, whereas for $\textit{BidiA}_{BLEU}$ and $\textit{BidiA}_{{WMD}_T}$ it varies. $\textit{BidiA}_{BLEU}$ that has a better BLEU-4 (see \autoref{fig:bleues}) than $\textit{BidiA}_{{WMD}_T}$ has a index distribution more similar to \textit{Best Hypothesis} than $\textit{BidiA}_{{WMD}_T}$. 

\subsection{Diversity of the responses}
\autoref{tab:div} has shown the performance in diversity metrics. Overall, \textit{BidiA} has the best performance among the other strategies (improvement up to 8\% over the baseline for Cornell). By looking for an agreement between the reverse seq2seq and the regular one \textit{BidiA} is able to extract sequences that are less likely according to the \textit{VBS}, but more diverse. In all case, we see that bidirectionally helps to have more diverse sentences. Since the influence of the input decreases during the generation bidirectional beam search will output sentences that have both meaningful beginning and ending with respect to the input.

\begin{table}

\centering
\begin{tabular}{ccc|cc}
\hline
&\multicolumn{4}{c}{distinct-n} \\\cmidrule(r){2-5} 
& \multicolumn{2}{c}{Cornell} & \multicolumn{2}{c}{Switchboard} \\
    Model &n=1 & n=2   &n=1 & n=2  \\ \hline
   \textit{VBS}  & 0.051 & 0.250   & 0.042 & 0.231  \\ 
   \textit{BidiS}  & 0.051 & 0.257  & 0.046 & 0.240  \\
   $\textit{BidiA}_{BLEU_T}$  &  \textbf{0.056}  & 0.261  & \textbf{0.050} & 0.240 \\
      $\textit{BidiA}_{WMD_T}$  & 0.054  & \textbf{0.270} & 0.048 & \textbf{0.241} \\
   \hline
 
\end{tabular}
\caption{\textbf{Diversity Scores} we report the diversity score (distinct-n) for $N_B=50$.}
\label{tab:div}
\end{table}


\subsection{Complexity Analysis}
In practical application it is important to evaluate the algorithm complexity when a limited amount of resources are available. \autoref{tab:open} shows that \textit{BidiA} is computationally cheaper than \textit{VBS} and that \textit{BidiS} has the same complexity as \textit{VBS}.
\begin{table}
\centering
\begin{tabular}{cc}
\hline
Algorithm & Complexity \\\hline
    $\mathcal{C}_{ \textit{VBS}}$  &$T \times \mathcal{O}( B  V \times \log(B V))$ \\ 
  $ \mathcal{C}_{\textit{BidiS}} $ & $ T \times \mathcal{O}(  B  V \times \log(BV))$  \\
 $  \mathcal{C}_{\textit{BidiA}_{WMD_T}} $ & $2 T \times \mathcal{O}(\frac{B}{2} V \times log(\frac{BV}{2})$ \\
  $  \mathcal{C}_{\textit{BidiA}_{BLEU_T}} $ & $2 T \times \mathcal{O}(\frac{B}{2} V \times log(\frac{BV}{2})$ \\
   \hline
\end{tabular}
\caption{\textbf{Complexity of the different algorithms.} V is the size of the dictionary, B is the beam size, T is the maximum sentence length.}
\label{tab:open}
\end{table}

\section{Conclusions}

\label{conclusions}
In this paper we show that bidirectional beam search strategies can be leverage to boost the performance of beam search. We have introduced two novel re-ranking criterions that select sentences with more diverse sentences and higher BLEU-4 and reduce computational complexity. Future work includes testing our novel bidirectional strategies with other pretrained models such as the one introduced in \cite{heavy_tailed,chapuis2021code,emile,classif}, with other types of data (\textit{e.g} multimodal \cite{garcia2019token,colombo2021improving}), on differents tasks (\textit{e.g} style transfer \cite{colombo2021novel}) as well as exploring other stoping criterions \cite{colombo2021automatic}.
\bibliography{naaclhlt2019}
\bibliographystyle{acl_natbib}
\newpage
\clearpage
\newpage
\section{Supplementary Material}
\subsection{Corpus analysis: importance of reverse model} \label{sub:reverse_data}
\begin{figure}
  \centering
  \includegraphics[width=0.48\textwidth]{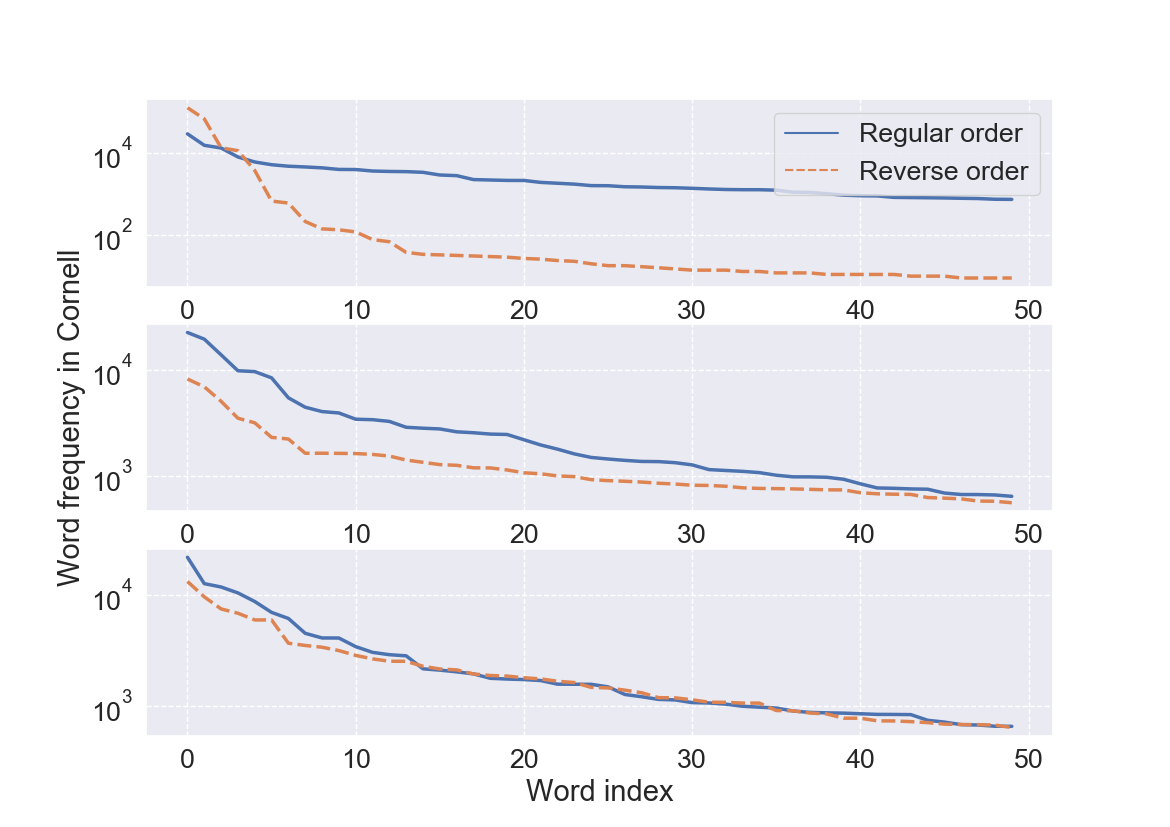}
  \caption{\textbf{Word distribution on Cornell for different word position in the sentence.} This figure shows the frequency of appearance of the 50 most common word in the sentence extracted from Cornell. An example of the regular order sentence is "The cat is red.", the associated reverse order is: ". red is cat The". Top plot shows the frequency of words appearing in position 1, the second plot for words appearing in position 2, the third plot for words appearing in position 3.}
  \label{fig:word_distribution}
\end{figure}

In this section, we discuss the importance of using a reverse model. \autoref{fig:word_distribution} shows the word distribution on Cornell for 50 most common words at each position. From top plot we observe that the regular seq2seq faces a lot of likely choices: all the fifty most common words appear more than $10^3$ times in position 1. The reverse seq2seq faces less than $5$ very likely choices in position 1 that appear more than $10^3$ time. The reverse seq2seq is then less likely to transmit a mistake at time step 2.

\subsection{Ideal reranking}\label{sub:ideal_reraking}

In \autoref{tab:faillure} we report the BLEU-4 achieved by \textit{VBS} and $\textit{Best Hypothesis}$: the best hypothesis alive in the beam. It illustrates that the limitations of the likelihood criterion and shows that making a change in the final reranking and sentence selection criterion can yield to higher BLEU-4. 
\begin{table}
\centering
\begin{tabular}{cccccc}
\cmidrule(r){2-6}
 && \multicolumn{4}{c}{BLEU-4}  \\\cmidrule(r){3-6}
 &Beam Size  &1  & 6 & 10 &   50\\\cmidrule(r){2-6}
  \parbox[t]{2mm}{\multirow{2}{*}{\rotatebox[origin=c]{90}{Corn.}}}   &$\textit{VBS}$  &1.19  & 1.23 & 1.30 &   1.30 \\ 
    & $\textit{Best Hypoth.}$  &1.19  & 1.40 & 1.60 &   2.56 \\  \cmidrule(r){2-6}
\parbox[t]{2mm}{\multirow{2}{*}{\rotatebox[origin=c]{90}{SWA}}}     & $\textit{VBS}$  &2.39  & 2.45 & 2.47 &   2.52 \\ 
     & $\textit{Best Hypoth.}$  &2.39  & 3.40 & 4.30 &   5.77 \\ 
\cmidrule(r){2-6}
\end{tabular}
\caption{\textbf{BLEU-4 Scores on Cornell (Corn.) and Switchboard (SWA)}: $\textit{VBS}$ stands for the standard beam search (see \autoref{models}) $\textit{Best Hypoth.}$ is the hypothesis in the beam that leads to the highest BLEU-4. In our work performances of $\textit{Best Hypoth.}$ can be seen as an upper bound of the performances of $\textit{VBS}$.}
\label{tab:faillure}
\end{table}

\subsection{Implementation of similarity measure ($sim$)}\label{sub:d_implem}
In this section we describes the implementation of each similarity $sim$ used in \autoref{tasks}. We have introduce a brevity penalty for two main reasons: 
\begin{itemize}
    \item preliminary experiments have shown that the regular seq2seq tends to generate short sentences due to the data distribution.
    \item if no brevity penalty is introduced and both neural networks generate ``I don`t know" the selected sentence will be ``I don`t know" since the similarity measure will be 1. With a brevity penalty, similarity metric can select a less generic choice.
\end{itemize}

\subsubsection{$BLEU_T$}$BLEU_T$ has been implemented by using the nltk librairy\url{https://www.nltk.org/}. In \autoref{eq:bidia} we set $sim = 1 -  BLEU_T$.
\subsubsection{$WM_T$}$WM_T$ uses the wm-relax librairy (\url{https://github.com/src-/wmd-relax}), embeddings used are coming from FastText librairy \cite{embeddings}. At the first step stopwords according nltk list are removed, Word Mover Distance is computed and multiplied by $BP_T$ previously defined. Formally, in \autoref{eq:bidia} we set $sim = WM_T$.

\subsection{Architecture details}\label{sub:archi_detail}
We evaluate our proposed algorithms by using off-the-shelf seq2seq models. For the encoder, we use two-layer bidirectional GRU \cite{gru} ($256$ hidden layers). For the decoder, we use a one-layer uni-directional {GRU} ($512$ hidden layers) with attention \cite{attention}. The embedding layer is initialized with fastText pre-trained word vectors (on Wikipedia 2017, the UMBC web-based corpus and the statmt.org news dataset) and the size is $300$ \cite{embeddings}. We use the ADAM optimizer \cite{adam} with a learning rate of $0.001$, which is updated by using a scheduler with a patience of $100$ epochs and a decrease rate of $0.5$. The gradient norm is clipped to $5.0$, weight decay is set to $1e^{-5}$, and dropout \cite{dropout} is set to $0.1$. The models have been implemented with pytorch, they have been trained on $97\%$, validated on $1\%$, and tested on $2\%$ of the data respectively. Since our purpose is to show that bidirectionality can boost beam search we set $\alpha = \frac{1}{2}$ in \autoref{eq:multitask}.

\subsection{Proofs of Complexity analysis}\label{sub:complexity}

\subsubsection{VBS complexity}
 For \textit{VBS}, at each time step $B V$, hypotheses are re-ranked and the $B$ most likely are kept. The final average complexity is:\vspace{-0.12cm}
\begin{equation}
\mathcal{C}_{ \textit{VBS}} = T \times \mathcal{O}( B  V \times \log(B V))
\vspace{-0.12cm}
\label{eq:vbs}
\end{equation}
\subsubsection{BidiS complexity}
In the case of \textit{BidiS}, the algorithm generates $B$ sequences using \textit{VBS}, and then for generating sequence $Y_{T}$ it computes $\frac{\log P(Y_T^-|X) }{lp(Y_T)}$ with complexity $\mathcal{O}(T)$. The final step includes a sorting of complexity $\mathcal{O}(B \log(B))$.\footnote{ $\mathcal{O}(B \log(B))$ and $\mathcal{O}(T)$ have much less order compared to $\mathcal{O}( B  V \times \log(B V))$  so they can be neglected here.} \textit{BidiS} complexity is:
\begin{equation}
\mathcal{C}_{ \textit{BidiS}}= T \times \mathcal{O}(  B  V \times \log(BV))†
\vspace{-0.12cm}
\end{equation}
\subsubsection{BidiA complexity}
\textbf{Word Mover's Distance criterion:} According to \cite{mover_distance} the computational cost of the Word Mover's Distance computation is $\mathcal{O}(p^3 \times \log(p))$, where $p$ denotes the number of unique words in the documents. In our case the distance is computed between two sequences of length at most $T$, hence $p \leq 2T$. $\textit{BidiA}_{WMD_T}$ complexity with Word Mover's Distance as selection criterion is given by the following formula:\vspace{-0.12cm}

\begin{equation}
\label{eq:bidia}
    \begin{split}
    \mathcal{C}_{ \textit{BidiA}_{WMD_T}} = & \underbrace{2 T \times \mathcal{O}(\frac{B}{2} V \times log(\frac{BV}{2})) }_{
\text{two \textit{VBS} with with beam size $\frac{B}{2}$}
} \\&+ \frac{B^{2}}{8} \times \underbrace{  \mathcal{O}(8 T^3 \times \log(2T))}_{\text{pairwise WMD}}\\& +  \underbrace{  \mathcal{O}(T)}_{\text{complexity of $BP_T$}}
\end{split}
\end{equation}
In general $T^3 \leq BV$ and $T \lll BV$, in Equation~\ref{eq:bidia} the second term is small compared to the first term, hence $\mathcal{C}_{ \textit{BidiA}_{WMD_T}} \approx  T \times \mathcal{O}(B V \times log(\frac{BV}{2}))$. Even thought V dominates the complexity of the algorithm, still $\textit{BidiA}_{WMD_T}$ is more efficient than \textit{VBS}. \footnote{For example if $T=30$, $B=30$, $V=35$k we see that $\mathcal{C}_{ \textit{VBS}} = 1.4 \times \mathcal{C}_{\textit{BidiA}}$.}

\noindent\textbf{BLEU criterion:} the computational cost of the $BLEU_T$ score is polynomial in T.  $\textit{BidiA}_{BLEU_T}$ complexity with BLEU score as the selection criterion is given by the following formula:\vspace{-0.12cm}

\begin{equation}
\label{eq:bidia}
    \begin{split}
    \mathcal{C}_{ \textit{BidiA}_{BLEU_T}} = & 2 T \times \mathcal{O}(\frac{B}{2} V \times log(\frac{BV}{2}))
\end{split}
\end{equation}
\end{document}